\pdfoutput=1

\documentclass[11pt]{article}

\usepackage{acl}

\usepackage{times}
\usepackage{latexsym}
\usepackage{enumitem}
\setlist[itemize]{noitemsep,align=parleft,left=0pt..1em,topsep=3pt}

\usepackage[T1]{fontenc}
\usepackage{todonotes}

\usepackage[utf8]{inputenc}
\usepackage[english, greek]{babel}
\usepackage{booktabs} 
\usepackage{multicol}
\usepackage{comment}

\usepackage{microtype}

\usepackage{inconsolata}
\usepackage{amsmath}
\usepackage{graphicx}
\usepackage[nameinlink,capitalize,noabbrev]{cleveref}

\title{Re-identification of De-identified Documents\\ with Autoregressive Infilling}

\author{Lucas Georges Gabriel Charpentier \\
  University of Oslo \\
  Language Technology Group \\
  \textsf{lgcharpe@ifi.uio.no} \\\And
  Pierre Lison \\
Norwegian Computing Center (NR), Oslo \\
\textsf{plison@nr.no}}

\begin{document}
\selectlanguage{english}
\renewcommand{\arraystretch}{1.0}
\maketitle
\begin{abstract}
Documents revealing sensitive information about individuals must typically be de-identified. This de-identification is often done by masking all mentions of personally identifiable information (PII), thereby making it more difficult to uncover the identity of the person(s) in question. To investigate the robustness of de-identification methods, we present a novel, RAG-inspired approach that attempts the reverse process of \textit{re-identification} based on a database of documents representing background knowledge. Given a text in which personal identifiers have been masked, the re-identification proceeds in two steps. A retriever first selects from the background knowledge passages deemed relevant for the re-identification. Those passages are then provided to an infilling model which seeks to infer the original content of each text span. This process is repeated until all masked spans are replaced. We evaluate the re-identification on three datasets (Wikipedia biographies, court rulings and clinical notes). Results show that (1) as many as 80\% of de-identified text spans can be successfully recovered and (2) the re-identification accuracy increases along with the level of background knowledge.
\end{abstract}

\section{Introduction}

 \begin{figure}[t]
     \centering
     \includegraphics[width=\linewidth]{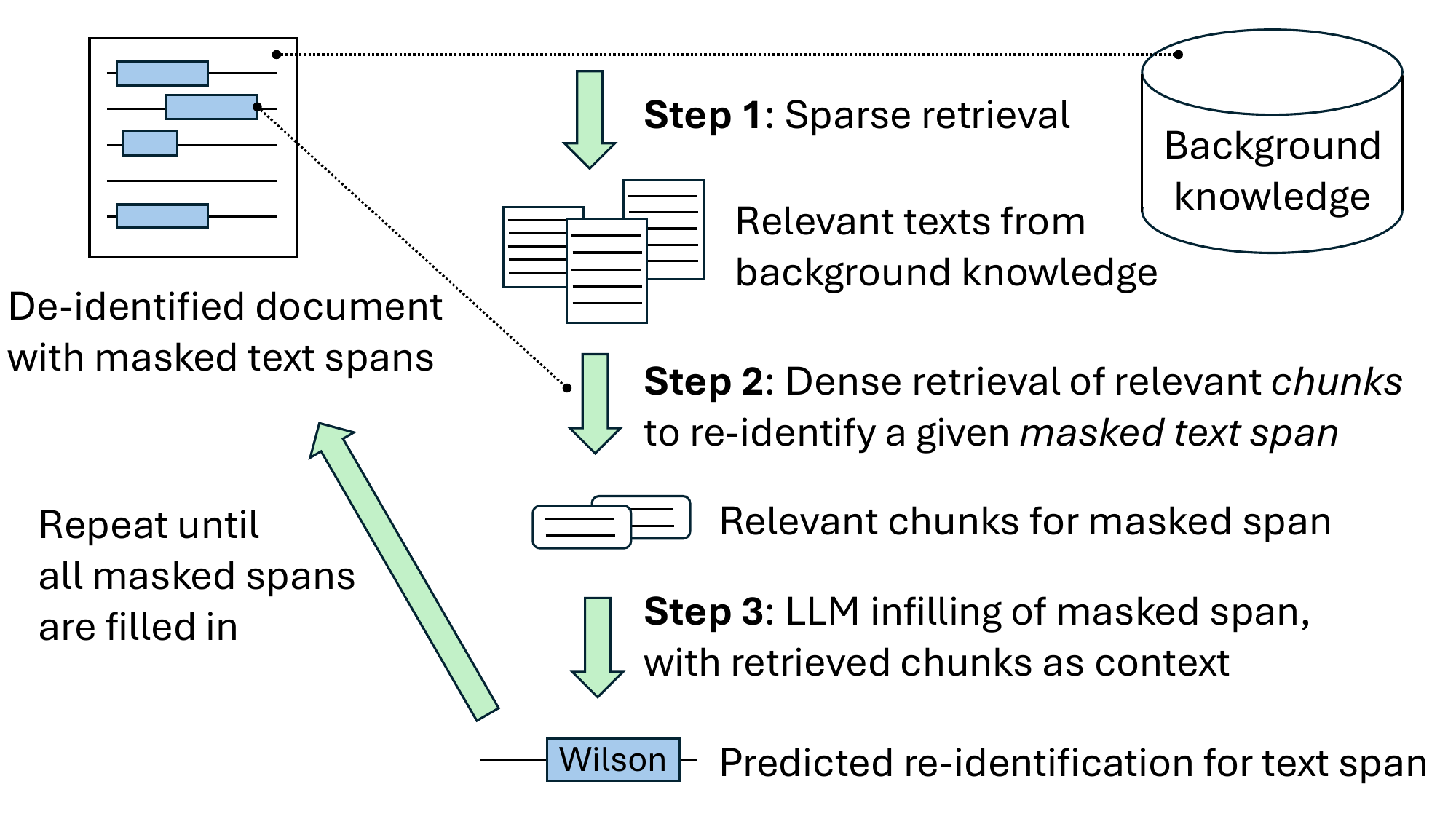}
     \caption{Sketch of the re-identification pipeline. The approach takes as input a document in which PII has been masked. A sparse retriever first selects relevant documents from the background knowledge. A dense retriever then extracts from those the chunks deemed most useful for re-identifying a particular text span. Finally, the infilling model produces a re-identification guess for that span given the retrieved chunks. The process is repeated until all text spans are filled back in.} \vspace{-2mm}
     \label{fig:enter-label}
 \end{figure}
 
Many types of text documents contain sensitive information about human individuals. When releasing or transferring those documents to third parties, it is typically desirable -- and often legally required -- to \textit{de-identify} them beforehand. Most de-identification approaches operate by (1) determining the text spans that express direct or indirect personal identifiers and (2) masking those from the document. This process can be done manually or using NLP models \cite{sweeney1996replacing,neamatullah2008automated,Sanchez2016,dernoncourt2017identification,lison-etal-2021-anonymisation,liu2023deid}.

It is, however, difficult to properly assess whether the de-identification has adequately concealed the identity of the person(s) mentioned in the original document. Many evaluation techniques assess the performance of de-identification methods by comparing their outputs with those of human experts \cite{lison-etal-2021-anonymisation,benchmark}. However, those evaluation techniques depend on the availability of human annotations and may be prone to human errors and inconsistencies.

An alternative approach to evaluating the robustness of the de-identification is through an automated {\it adversary} carrying out re-identification attacks \cite{manzanares2024evaluating}. This paper presents such an adversarial approach, based on a retrieval-augmented scheme where relevant information is first retrieved from a body of background knowledge, and then exploited to infer the original content of each masked text span. The background knowledge should represent all information that one assumes may be available to adversaries. As shown by the evaluation results, the amount of information included as background knowledge notably influences the re-identification accuracy. 

\cref{sec:back} provides a short background on text de-identification, text infilling and retrieval augmentation. \cref{sec:methods} describes the re-identification approach, which is then evaluated in \cref{sec:evaluation} on three datasets. Finally, Sections \ref{sec:disc} and \ref{sec:con} discuss the results and outline future directions.

 \section{Background} \label{sec:back}

 \subsection{Text de-identification}
 
 The processing of personal data is regulated through legal frameworks such as the European \textit{General Data Protection Regulation} \cite{gdpr}. A key principle of those frameworks is \textit{data minimization}, which states that data owners should restrict data collection and processing to only what is required to fulfill a specific purpose. The goal of text de-identification, also called text sanitization \cite{Sanchez2016,papadopoulou-etal-2022-neural}, is precisely to enforce this minimization principle by concealing personal identifiers from the text \citep{lison-etal-2021-anonymisation, benchmark}. 

Personally identifiable information, or PII, can be divided into two categories, both of which should be masked from the text to ensure the texts are properly de-identified \citep{soton399692}:
\begin{itemize}
     \item \textbf{Direct identifiers}, which relate to information that can univocally identify a person, such as their name, phone number or home address. 
    \item \textbf{Quasi identifiers}, which are not specific enough to single out an individual alone, but may do so when combined together. Examples include the person's nationality, gender, occupation, place of work, date of birth or physical appearance.
\end{itemize}

Evaluating de-identification methods is a challenging task. A common solution is to compare the masking decisions of the model against manual annotations \citep{benchmark}. Such reference-based evaluations are, however, not always feasible, and are hampered by residual errors, omissions, and inconsistencies in human judgments. 

One alternative is to carry out re-identification attacks on the de-identified documents to determine whether an adversary is able to uncover the identity of the person to protect \citep{scaiano2016unified,mozes2021no}. \citet{morris-etal-2022-unsupervised} present a model for inferring infoboxes from a sanitized Wikipedia page. This model is employed to guide the masking choices of a text de-identifier such that the correct infobox can no longer be predicted from the edited text. This approach was recently extended to the medical domain in \citep{morris2024diriadversarialpatientreidentification}. Contrary to this paper, they do not attempt to re-identify the masked spans themselves. \citet{manzanares2024evaluating} train a neural text classifier to link back Wikipedia biographies with its corresponding person name. This classifier, however, directly predicts the person's name from the text. In contrast, the approach present in this paper takes advantage of LLMs to first uncover the masked text spans and only seeks to predict the person's identity after this unmasking step. 

The idea of building an adversary to unveil a sensitive attribute has also been explored in the area of text rewriting \cite{xu-etal-2019-privacy}. However, those approaches typically seek to protect other attributes than the person's identity (such as gender or ethnicity) and focus on different types of document edits than PII masks. Such complete rewrites of the text can also performed using methods based on differential privacy \citep{igamberdiev2023dpbart}, although those methods do not typically conduct explicit re-identification attempts.
 
 \subsection{Text infilling}

The prediction of missing/masked spans of text at any position within a document (often indicated via a special placeholder symbol) is known as \textit{infilling} \citep{infill,donahue-etal-2020-enabling} or {\it fill-in-the-middle} \citep{bavarian2022efficient}. In contrast to masked language models such as BERT \citep{devlin-etal-2019-bert}, which are pretrained to infer a single masked token based on the surrounding context, the infilling task may span multiple tokens (whose number is typically left unknown, although one can control its length).  Two early approaches to text infilling were respectively presented by \citet{infill} and \citet{donahue-etal-2020-enabling}. Those two approaches demonstrated how to pre-train and fine-tune a language model to fill in spans of a controlled size. More recently, a Generalized Language Model (GLM) was proposed by \citet{du-etal-2022-glm}, unifying both encoder and decoder architectures. GLM can be seen as generalizing the token-level masking of encoder models by (1) masking entire spans with a single token and (2) training the model to autoregressively generate the correct replacement span at the end of the text. 


\subsection{Retrieval-augmented models}

The factual knowledge stored in LLMs is distributed among all parameters and cannot be easily edited, updated, or inspected.  \textit{Retrieval-augmented language models} \citep{NEURIPS2020_6b493230,10.5555/3524938.3525306,ram-etal-2023-context} address this shortcoming by coupling the model with a knowledge base of documents. The generation process is then split into a \textit{retrieval} phase, in which relevant documents from the knowledge base are extracted, and a \textit{reading} phase, which corresponds to the actual generation, conditioned on both the context and the relevant documents selected by the retriever. 

Retrieval-Augmented Generation (RAG) make it possible to edit or update the knowledge base while leaving the underlying model unchanged \citep{gao2023retrieval}. The retrieval mechanism can also enhance the system's interpretability, as one can inspect the retrieved documents and assess their influence in the final output \citep{sudhi2024rag}. 

Retrieval-augmented models can be trained in multiple ways. A common strategy is to start with pre-trained retriever and reader models, and fine-tune those two end-to-end on a standard language modelling objective \citep{NEURIPS2020_6b493230}. One can also continue the model pre-training with a retriever that can be trained \citep{10.5555/3524938.3525306} or not \citep{izacard2023atlas}. Models trained from scratch with a trained retriever have also been proposed \citep{borgeaud2022improvinglanguagemodelsretrieving}.

The approach described in this paper is directly inspired by RAG architectures, as we also rely on a neural retriever connected with a knowledge base. However, while most previous work on RAG has concentrated on tasks such as question answering, we focus here on the task of re-identifying a document in which PII have been masked. 

 \section{Approach} \label{sec:methods}

The proposed method is divided into three steps, as illustrated in Figure \ref{fig:enter-label}. A \textit{sparse retriever} is first employed to find relevant background documents for the de-identified text. For each masked text span of the de-identified text, we then perform a \textit{dense retrieval} to determine the passages in the selected background documents that are most relevant to unmask that span. Using those passages, a fine-tuned LLM then generates \textit{infilling hypotheses} for the masked span. The operation is repeated until all masked spans in the de-identified document are replaced by their most likely hypothesis. 

\subsection{Sparse document retriever}

The sparse retriever takes as input a de-identified text and outputs a list of relevant background documents. Those documents are retrieved from a large database which should ideally comprise all information that one can expect to be available to an adversary seeking to uncover the personal information that the de-identification sought to conceal. 

To efficiently search for those documents, we rely on the BM$\mathcal{X}$ algorithm \citep{li2024bmxentropyweightedsimilaritysemanticenhanced}, a modified version of the BM25 algorithm \citep{robertson2009probabilistic} which takes into account lexical and semantical similarities, with a default setup and retrieve the $N$ most similar documents (where $N$ was set to 100 in our experiments).


\subsection{Dense passage retriever}

The documents selected by the sparse BM$\mathcal{X}$ retriever are then split into overlapping chunks of about 600 characters each. For each masked span in the de-identified document, we create a query string of 128 tokens consisting of the local context around that span. The masked span in that query is denoted with a special \texttt{[MASK]} token. 

The dense retriever is a fine-tuned ColBERT \citep{khattab2020colbert}. The data employed for the fine-tuning consists of both positive and negative (passage, query) pairs. The positive pairs are defined as passages that include the original content of the span that was masked, while the negative pairs are passages that do not. For instance, if the sentence ``The applicant lives in the German city of Aachen'' was de-identified as ``The applicant lives the German city of \texttt{[MASK]}'', the pair $\langle$``Aachen is the westernmost city in Germany', ``The applicant was born in the German city of \texttt{[MASK]}''$\rangle$ will constitute a positive example for the retriever. This setup makes it possible to fine-tune the ColBERT retriever independently of the infilling model. 


\subsection{Infilling}

The passages selected by the retriever are then used to produce re-identification candidates for each masked span in the input document. Next to those passages, we also provide the context of the masked span in the document, such as ``The applicant lives in the German city of \texttt{[MASK]}''.  

We experiment with two distinct LMs to generate infilling hypotheses. The first is a GLM RoBERTA Large \citep{du-etal-2022-glm}, where the context is provided with a 200-character window to the left and the right of the span. 
While we could in principle use the GLM to generate hypotheses without fine-tuning, we found that fine-tuning improved the infilling results, as it incites the LLM to exploit the information in the retrieved passages in addition to the span context.

We also run an instruction-tuned version of Mistral-12B, Mistral-Nemo-Instruct-2407\footnote{\url{https://huggingface.co/mistralai/Mistral-Nemo-Instruct-2407}} ( \citep{jiang2023mistral7b}) with the same context as for the GLM.
The model is run without shots due to the difficulty of providing suitable examples of re-identification with associated retrieval passages within the limits of the context window. 


Given a de-identified document, we replace each masked span one at a time, in randomized order, until all masked spans are replaced. 

\section{Evaluation} \label{sec:evaluation}

We evaluate the approach on three datasets. The first one is a generic corpus extracted from Wikipedia in which personal identifiers have been masked with a standard Named Entity recognizer. The second is the Text Anonymization Benchmark \citep{benchmark}, which was explicitly designed for privacy-oriented NLP tasks, and has been manually annotated with both direct and quasi-identifiers. The final dataset a set of synthetic clinical notes generated from (also synthetic) patient records. 

To assess how background knowledge influences re-identification, we conduct the evaluation with four levels of background knowledge:
\begin{itemize}
\item \textbf{L1 - No retrieval}: No background knowledge is assumed and the infilling is performed directly, without including any retrieved passage.
\item \textbf{L2 - General knowledge}: We include a set of background documents, but without the original version of the texts that were de-identified. 
\item \textbf{L3 - All texts except input document}: This setup extends the general knowledge with the original version of the de-identified texts, except the one we currently seek to re-identify.
\item \textbf{L4 - All texts, including input document}: This setup mimics a strong adversary with  access to background documents including the original version of the text to re-identify.
\end{itemize}

\subsection{Data}
\label{sec:data}

 \subsubsection*{Wikipedia Biographies}

    The Wikipedia biographies consists of all English-language biographies identified by the Biography WikiProject.\footnote{\url{https://en.wikipedia.org/wiki/Wikipedia:WikiProject_Biography}}, amounting to over 2M biographies. This dataset is used to fine-tune both the retriever and infilling model. 
    We de-identify the biographies by running an English NER model from Spacy\footnote{\url{https://spacy.io/models/en\#en_core_web_trf}} and masking every detected entity\footnote{Although not all named entities are personal identifiers, and personal identifiers may also correspond to expressions that are not named entities, there is a strong correlation between the two, especially in Wikipedia biographies.}.

    To define the general background knowledge (L2), we use the rest of English-language Wikipedia (i.e. all non-biographies) which represents about 4.7M articles. These articles could relate to e.g.~discoveries or events connected to the person described in the biography. For levels 3 and 4, we also include the Wikipedia biographies themselves, respectively without and with the actual biography to re-identify.

    Due to the large size of this dataset, we only use the GLM to infill the masked spans of those biographies, while both GLM and Mistral are used for the two other datasets.

 \subsubsection*{Text Anonymization Benchmark (TAB)}

    The TAB dataset \citep{benchmark} consists of 1\,268 English-language court cases from the European Court of Human Rights (ECHR). 
    Each court case has been manually de-identified and includes detailed annotations such as identifier type, semantic category and confidential attributes.

    Level 2 of background knowledge is compiled from 28\,569 legal summaries, reports, and communicated cases from the ECHR. To further increase the volume of background knowledge, we also include three generated articles (a news article, a blog post and a court report) using Mistral-Nemo-Instruct-2407 for each of the test cases. The prompts for those generations can be found in \cref{app:prompts}. Levels 3 and 4 also include the court cases themselves as well as the court cases from the train set of the TAB corpus.

\subsubsection*{Synthetic Clinical Notes}

    The clinical notes consist of 1-10 patient notes for 85 distinct patients, resulting in 298 patient notes. Patient records were first generated using the Synthea patient population simulator \citep{walonoski2018synthea}. Clinical notes were then generated from the resulting patient records with a dedicated, GPT4.0 powered tool provided together with the Synthea simulator\footnote{\url{https://github.com/synthetichealth/chatty-notes}}. The notes are de-identified in the same way as the Wikipedia Biographies.
        
    Level 2 of background knowledge consists of 1\,146 synthetic records (including the 85 patients described in the notes) in YAML format.  Levels 3 and 4 also include the original notes themselves, either without or with the note to re-identify.

\subsection{Training details}
\label{sec:training_details}
\subsubsection*{Retrieval}

We train the ColBERT model for the dense retrieval with the de-identified Wikipedia biographies and the non-biographies as databases. After splitting the background documents in chunks, we create a training set with $\langle$ passage, query $\rangle$ pairs consisting of both positive examples containing the span content and negative examples that do not contain it. To increase the pool of positive examples, we use Wikipedia re-directions to get alternative spellings of the span content, such as viewing ``J.F.K.'' as equivalent to ``John F. Kennedy''\footnote{We obtain those Wikipedia redirections from  \url{https://github.com/Social-Data-inSights/coproximity_create_vocabulary} }.  

We fine-tune the ColBERT retriever for English, more precisely two case-sensitive base-sized BERTs for respectively embedding the documents and queries. We train the retriever on 127K training examples for 20K steps with a batch size of 256 and a learning rate of $3\times10^{-5}$ and compress each document and query token from 768 dimensions down to 32. As in \citep{khattab2020colbert}, we fix the sequence length of the queries to 128 tokens and use the extra tokens as "memory tokens" to embed extra information to help find relevant documents. For the train set, we only consider spans having at least two positive chunks since there is always one from the original document.



\subsubsection*{Infilling}

We train the infilling model with a dataset consisting of de-identified chunks from Wikipedia biographies and their top ColBERT retrieved text, amounting to about 160K training examples. We train them for one epoch with a batch size of 128 and a learning rate of $3\times10^{-5}$, where each data point is distinct (i.e. there is no repeated training sample). 
Due to context window limits for the GLM, we only include 1 or 2 retrieved passages in the input for this model, while the Mistral model uses the top-10 retrieved passages. 


All models are trained with a single GPU (RTX3090 for ColBERT, A100 for the infilling models). In total, the training took 10 hours. Inference per run takes about 30 minutes for the GLM and 1h30 for Mistral.

\subsection{Final re-identification}
\label{sec:final_re_id}

We also test whether the re-identification approach can be employed to determine the exact name of the person the text relates to. This is done by training a ranking model that takes as input (1) the document infilled by the model with re-identification candidates and (2) a list of $N$ candidates, such as the names of all persons known to appear in a given dataset. This ranking model relies on a BERT model fine-tuned with a margin ranking loss objective. 
We train on a dataset of 2.3K infilled documents, with a batch size of 32, a learning rate of $3\times10^{-6}$, and for 40 epochs.

\subsection{Metrics} \label{sec:metrics}

For testing, we respectively used 298 held-out Wikipedia biographies, the test set of the TAB corpus (127 court cases), and 298 patient notes from 85 patients. 

We first analyze the performance of the sparse and dense retrievers, and then evaluate the end-to-end performance of the complete system.

    \paragraph{Sparse Retrieval}

To evaluate the performance of the sparse retrieval mechanism, we look at the percentage of masked spans in a sanitized text that can be found in the top 100 retrieved documents. 
    
    \paragraph{Dense Retrieval}

        We use both Mean Reciprocal Rank (MRR) and accuracy@k (specifically @1, 5, and 10) to assess the dense retrieval accuracy. If the retrieved text has the span to re-identify, it is considered a positive instance. However, given not all spans have a retrieved chunk with a correct answer, we only look at spans where the masked span exists in one of the retrieved chunks. 


    \paragraph{Infilling}

        We use two metrics to judge the accuracy and performance of our re-identifications. The first is an exact match, where a re-identification is only correct if it outputs the original tokens. The second is token recall where we look at the proportion of tokens in the prediction that are also in the original span. This token recall makes it possible to give partial credit to shorter names that refer to the same entity (i.e.~``President Emmanuel Macron'' and ``Macron''). 

        


    \paragraph{Final re-identification}

        For the problem of ranking candidate names for the person whose identity was concealed in the document, we provide results for accuracy@10. The results for accurary@1,@5 and MRR are detailed in the appendix.

\subsection{Results}\label{sec:exp}

 \begin{table}[t]
     \centering
     \small
     \begin{tabular}{@{}lccc@{}} 
          \toprule  
          \textbf{Dataset} & \textbf{General} & \textbf{All but original} & \textbf{All} \\
          \midrule
          Wikipedia & 53.4\textsuperscript{$\pm{16.7}$} & 60.2\textsuperscript{$\pm{16.8}$} & 98.2\textsuperscript{$\pm{10.2}$} \\
          TAB & 64.5\textsuperscript{$\pm{18.0}$} & 75.1\textsuperscript{$\pm{14.6}$} & 100\textsuperscript{$\pm{0.0}$} \\
          Clinical & 51.0\textsuperscript{$\pm{15.8}$} & 88.2\textsuperscript{$\pm{13.3}$} & 99.8\textsuperscript{$\pm{1.2}$} \\
          \bottomrule
     \end{tabular}
     \caption{Percentage of masked spans from the de-identified test documents found in the top-100 documents of the sparse retrieval.}
     \label{tab:bm25} \vspace{-1mm}
 \end{table}

\begin{table}[t]
     \centering
     \small
     \begin{tabular}{@{}lcccc@{}} 
          \toprule  
          \textbf{Knowledge} & \textbf{MRR} & \textbf{Acc@1} & \textbf{Acc@5} & \textbf{Acc@10} \\
          \midrule
          \textsc{Wikipedia} & & & & \\
          \hspace{.3em}Not biographies & 0.175 & 10.6 & 22.1 & 27.3 \\
          \hspace{.3em}All but original & 0.229 & 14.0 & 30.5 & 40.4 \\
          \hspace{.3em}All & 0.895 & 87.8 & 91.5 & 94.0 \\[0.5em]
          \textsc{TAB} & & & & \\
          \hspace{.3em}General & 0.454 & 37.0 & 53.4 & 64.3 \\
          \hspace{.3em}All but original & 0.449 & 38.0 & 50.3 & 59.0 \\
          \hspace{.3em}All & 0.910 & 86.3 & 94.7 & 98.8 \\[0.5em]
          \textsc{Clinical} & & & & \\
          \hspace{.3em}General & 0.958 & 94.3 & 97.3 & 97.3 \\
          \hspace{.3em}All but original & 0.660 & 53.7 & 83.4 & 87.2 \\
          \hspace{.3em}All & 0.956 & 92.6 & 99.3 & 99.6 \\
          \bottomrule
     \end{tabular}
     \caption{Performance of the ColBERT dense retriever on spans with an existing retrieved chunk from the top-100 documents selected by the sparse retriever. The results are obtained on fully de-identified texts.}
     \label{tab:ColBERT} \vspace{-2mm}
 \end{table}

\subsubsection{Sparse and Dense Retrieval}


 We first analyse the performance of the sparse and dense retrieval steps. \cref{tab:bm25} details the percentage of masked spans from the test documents which were found in the top 100 documents selected by the sparse retriever across the three datasets. This performance increases along with the level of background knowledge, but we have high variations between biographies (around 15\%). This is possibly due to the notoriety of the person in the biography. The more notable a person is, the more likely non-biography texts will contain information on the person. Once we include the original biography, the results jump to 98.2\%. While this difference is high, it is expected, as the original version of the document to re-identify is here included as part of the background knowledge.

The sparse retrieval step is easier for TAB and the clinical notes than for the Wikipedia biographies. 
This is likely due to the smaller background for those two datasets. 
Reaching L4, the performance of the sparse retriever is either perfect (TAB) or near-perfect (clinical notes). 


 \cref{tab:ColBERT} shows a similar trend for the dense retriever where the performance increases along with the level of background knowledge. The results stay relatively low for Wikipedia in L2 and L3 (under 15\% for accuracy@1). As mentioned before, we only consider masked spans found in documents retrieved by the sparse retriever. Once we include the original text, the performance substantially increases (reaching 87.8\% for accuracy@1). 

We see that the performance of the ColBERT model (which was, as explained in Section \ref{sec:training_details}, fine-tuned on Wikipedia biographies) performs better on the TAB dataset than on the Wikipedia biographies at all levels of background knowledge. This could be due to the structured style of writing found in court cases. Both the accuracy and MRR increase for TAB along with the levels of background knowledge, reaching up to, for L4, 86.3\% of masked spans appearing in the top document retrieved by our ColBERT. 

The same trends can be found in the retrieval of the clinical notes. For the ColBERT retriever, we have higher performance for L2 and L3 compared to the previous datasets. However, both the texts and background knowledge are comparatively smaller, leading to fewer chunks in total as well as very structured and similar notes.

 \subsubsection{End-to-end infilling}

 \begin{table*}[ht]
    \small
     \centering
     \begin{tabular}{lccccccc@{}} 
          \toprule  
          \multicolumn{1}{@{}c}{\textbf{NER Category}}&  \textbf{No retrieval (L1)}&  \multicolumn{2}{c}{\textbf{Not Biographies (L2)}}&  \multicolumn{2}{c}{\textbf{All but not original (L3)}}& \multicolumn{2}{c}{\textbf{All (L4)}}\\
          \cmidrule(lr){3-4}\cmidrule(lr){5-6}\cmidrule(l){7-8}
          & & k=1 & k=2 & k=1 & k=2 & k=1 & k=2 \\
          \midrule
        Exact Match & \hphantom{0}6.26 & \hphantom{0}7.63 & \hphantom{0}7.71 & \hphantom{0}9.56 & \hphantom{0}9.77 & 80.08 & 78.99 \\
        Token Recall & 12.22 & 13.80 & 13.81 & 15.84 & 16.05 & 82.56 & 81.67 \\
         \bottomrule
     \end{tabular}
     \caption{Results of the GLM infilling at multiple background knowledge levels and numbers of retrieval texts on the Wikipedia biographies The overall results are bolded. The results represent the averages of 3 different runs, the standard deviation is less than 1\%. More details can be found in \cref{app:glm_recall}.}
     \label{tab:glm_wiki_re_id} \vspace{-1mm}
 \end{table*}
 
 \begin{table*}[ht]
     \centering
     \small
     \begin{tabular}{lrrrlrlrr} 
          \toprule  
          \multicolumn{1}{@{}c}{\textbf{Entity Category \ }} &  \multicolumn{2}{c}{\textbf{No retrieval  (L1) \ }} &  \multicolumn{2}{c}{\textbf{General Knowledge (L2) \  }} & \multicolumn{2}{c}{\textbf{All but not original (L3)  \  }} & \multicolumn{2}{c}{\textbf{All (L4)}} \\
          \midrule
            \multicolumn{1}{@{}l}{\textsc{GLM}} & 0.84 & 6.26 & \hskip15pt\relax 11.27 & 21.35 \hskip15pt\null & \hskip15pt\relax 14.32 & 29.08 \hskip15pt\null & 66.04 & 75.13 \\
            DIRECT & 0.00 & 6.61 & 0.82 & 7.12 & 0.81 & 15.23 & 28.60 & 47.63 \\
            QUASI & 0.90 & 6.23 & 12.08 & 22.63 & 15.40 & 30.54 & 68.93 & 77.82 \\
            \midrule
            \multicolumn{1}{@{}l}{\textsc{Mistral}} & 0.91 & 25.36 & 10.59 & 47.43 & 11.00 & 47.98 & 37.34 & 70.29 \\
            DIRECT & 0.00 & 15.84 & 5.71 & 36.36 & 0.90 & 33.96 & 15.39 & 48.86 \\
            QUASI & 0.98 & 25.90 & 10.97 & 48.49 & 12.12 & 49.25 & 39.03 & 72.21 \\
          \bottomrule
     \end{tabular}
     \caption{Results of infilling at multiple background knowledge levels on TAB. The first result represents exact match performance and the second is token recall. All results are the average of 3 runs, the standard deviation is less than 1\%. Detailed results can be found in \cref{app:glm_recall,app:mistral}.}
     \label{tab:tab_re_id} \vspace{-1mm}
 \end{table*}

 \begin{table*}[h!]
     \centering
     \small
     \begin{tabular}{lrrrlrlrr} 
          \toprule  
          \multicolumn{1}{@{}c}{\textbf{Entity Category} \ } &  \multicolumn{2}{c}{\textbf{No retrieval (L1) \ }} &  \multicolumn{2}{c}{\textbf{General Knowledge (L2) \ }} & \multicolumn{2}{c}{\textbf{All but not original  (L3) \ }} & \multicolumn{2}{c}{\textbf{All  (L4)}} \\
          \midrule
            \multicolumn{1}{@{}l}{\textsc{GLM}} & 18.31 & 26.71 & \hskip15pt\relax 18.92 & 26.36 \hskip15pt\null & \hskip15pt\relax 42.31 & 55.40 \hskip15pt\null & 90.87 & 92.68 \\
            \multicolumn{1}{@{}l}{\textsc{Mistral}} & 4.76 & 23.46 & 4.52 & 17.08 & 19.88 & 40.80 & 30.19 & 57.85 \\
          \bottomrule
     \end{tabular}
     \caption{Results of infilling at multiple background knowledge levels on the Clinical Notes Dataset. The first result represents exact match performance and the second is token recall. All results are the average of 3 runs, the standard deviation is less than 1\%. Detailed results can be found in \cref{app:glm_recall,app:mistral}.}
     \label{tab:clinical_re_id} \vspace{-1mm}
 \end{table*}
 


\paragraph{Wikipedia biographies}
\cref{tab:glm_wiki_re_id} details the end-to-end re-identification performance (in terms of exact match and token recall) for the Wikipedia biographies, using the GLM. The re-identification accuracy increases together with the level of background knowledge, with a small increase between L1 to L3 and a big jump once the original text is included in L4. It is consistent through nearly all NER Categories (see \cref{app:glm_recall}). 
The use of two retrieved passages instead of one leads in most cases to a small increase in performance, although at the cost of a substantial increase in compute time (around 40\% increase in sequence length). 

\paragraph{TAB}
\cref{tab:tab_re_id} shows the end-to-end re-identification performance on the TAB dataset for both the GLM and Mistral models. For the GLM, we used only one retrieved text for re-identification since the gains from using two were minor. For Mistral, we use the top 10 retrieved passages. Again, we observe that using any level of background knowledge is beneficial. Without retrieval, the model cannot re-identify any direct identifiers while each additional level of background knowledge leads to a small increase in the re-identification performance (measured in exact match and token recall). Once we include the original court case in the background knowledge (L4), the exact match for direct identifiers jumps to 28.6\%. For quasi-identifiers, we have the same trend of increasing performance as the background knowledge increases. \cref{tab:glm_tab_re_id} in the Appendix factors those results by entity categories. 

We observe the same trends for Mistral as in the GLM results. The token recall is, however, much higher for Mistral when using background knowledge L1 to 3, while L4 performs worse (in both exact match and token recall). While results for L1-L3 are expected (given the larger model size), the results for L4 are somewhat surprising and seem to indicate that the Mistral model gets confused in the L4 setup. In addition, the high token recall indicates that the model outputs reformulations of the span to re-identify. A manual analysis of the infilling outputs shows that the Mistral model tends to over-predict dates and numbers rather than codes and names. Examples of re-identifications can be found in \cref{app:re-identification}.


\paragraph{Clinical notes}
 \cref{tab:clinical_re_id} provides the re-identification results for the clinical notes. We see two major differences compared to the two previous datasets. The first is that the gap in performance between L1 and L2 (in which the background knowledge corresponds to the patient records in YAML format) is almost nonexistent. This might be because the model struggles to retrieve relevant information from the YAML-encoded patient records. The second is that the performance of the GLM is much better than Mistral for L3 and L4 (reaching over 90\% exact match on L4). This might indicate that the Mistral model fails to grasp which passage is most relevant due to similarities between patient notes. 

 \subsubsection{Final re-identification}

\begin{table}[ht]
     \centering
     \small
     \begin{tabular}{@{}lccccc@{}} 
          \toprule  
          \textbf{Dataset} & \textbf{Masked} & \textbf{L1} & \textbf{L2} & \textbf{L3} & \textbf{L4} \\
          \midrule
          \textsc{GLM} & & & & & \\
          \hspace{.3em}TAB & 28.3 & 32.3 & 31.5 & 29.1 & 61.4 \\
          \hspace{.3em}Clinical & 57.0 & 62.4 & 62.1 & 77.9 & 98.7 \\[0.5em]
          \textsc{Mistral} & & & & & \\
          \hspace{.3em}TAB & 28.3 & 32.3 & 33.1 & 37.0 & 57.5 \\
          \hspace{.3em}Clinical & 57.0 & 61.1 & 66.1 & 81.2 & 97.0 \\
          \bottomrule
     \end{tabular}
     \caption{Percentage of re-identified documents in which the correct identity is found in the top-10 predictions. More results can be found in \cref{app:re-identification}.}
     \label{tab:ranking} \vspace{-2mm}
 \end{table}
 
 \cref{tab:ranking} shows the results of the last experiment in which a BERT-based ranking model is employed to predict the exact identity of the person the document relates to (as explained in Section \ref{sec:final_re_id}). The number of candidates considered is equal to the number of cases (127) for TAB and the number of patients (85) for the clinical notes. For TAB, the table shows that the risk of singling out the exact identity of the person remains small unless the original document can be found in the background knowledge. 
 The ability of the ranking model to single out the person's identity is notably better for the clinical notes. However, the synthetic nature of those notes may have introduced artefacts enabling the ranking model to figure out the identify of the person from other textual cues. 
 
 The overall low increase in the re-identification accuracy as a function of the infilling performance seems to indicate that further background information must be provided to the ranking model to gain better accuracy at this singling-out task. 

 \section{Discussion} \label{sec:disc}

    Overall, we observe that having background knowledge closely related to the text or spans to re-identify leads to better re-identification of the spans. Usually, unique or uncommon categories of spans (such as direct identifiers) are harder to re-identify than more common ones (such as location, numbers, or demographics). We also notice that using the top retrieved document gives a big performance boost while adding a second retrieved document leads to minor improvements at a high cost. 

    The results obtained with Mistral indicate that the infilling step can be achieved with an instruction-tuned LLM without domain-specific fine-tuning or in-context demonstrations. This is encouraging, as the process of fine-tuning an LLM on domain-specific data for this infilling task may carry risks of privacy leakages \citep{kim2024propile}.  
    
We also observed that retrievers with architectures originally designed for QA-oriented retrieval problems, where named entities play a big part in the retrieval, still perform relatively well on the slightly different task of finding the most useful passages to re-identify spans. 

    Finally, we see that the risk of singling out the exact name of the person mentioned or referred to in the document remains relatively small unless we assume a strong adversary with access to background knowledge including the original version of the document to re-identify (L4) or who can restrain the list of candidates to a relatively small set of person names (as in the clinical notes). 

 \section{Conclusion} \label{sec:con}

    This paper presented a novel approach to the task of \textit{re-identifying} text documents that had previously been de-identified through the masking of personal identifiers. Automated re-identification models constitute an important tool for enhancing the robustness of current text de-identification methods, in particular to establish whether the content of a masked text span can be inferred from the context and available background knowledge.

    The presented method relies on a retrieval-augmented architecture that comprises a sparse retriever, a dense retriever, and an infilling model that takes advantage of the passages extracted in the retrieval phase. The method is evaluated on 4 distinct levels of background knowledge, and using three datasets: Wikipedia biographies, the Text Anonymization Benchmark (TAB), and a collection of synthetic clinical notes. We observed that texts de-identified either through NER (Wikipedia biographies and the clinical notes) or manually (TAB) can be at least partly re-identified. However, the re-identification performance strongly depends on the background knowledge assumed to be available to an adversary. Furthermore, even for modest levels of background knowledge, most quasi-identifiers can be correctly re-identified.

    
    Future work will extend the approach in several directions. The dense retriever is currently fine-tuned with a dataset of positive/relevant documents, where a document is deemed relevant if it contains the original string of the masked span (or one of its spelling variants). However, this has a number of shortcomings such as retrieving texts containing the correct span but in irrelevant contexts (There were \emph{four} objects considered vs. he had won \emph{four} gold medals), or texts that do not contain the correct span but could still provide useful information for the infilling. Improving the fine-tuning of this retriever could lead to better downstream results. Extending the background knowledge with other types of information (such as information derived from structured databases) could also enhance the re-identification performance. 



 \section*{Limitations}

 We only looked at texts in the English language and only used text data to help the re-identification, it is possible that using other types of data such as tables or knowledge graphs could be more helpful to this task. In addition, our infilling model, does not have a large amount of variant in different languages. However, looking at other languages might change our results. Also, our fine-tuned infilling model is relatively small (335M) and we only test the Mistral-12B model in a zero-shot manner rather than with In-context Learning. This is due to some computing constraints. Using In-context Learning or fine-tuning larger models could result in better re-identification due to better pattern matching. In addition, the Wikipedia and court rulings from TAB originate from text sources which are otherwise available on the web in clear text. This means that there is a possibility that some of the data has been leaked to the infilling model during the pre-training of it, thereby inflating the re-identification performance compared to documents which do not have a public, non-de-identified version available online.

 \section*{Ethical Statement}

 We acknowledge that creating models to re-identify sanitized texts could help attackers re-identify private data. However, our goal with this paper is to show that if it is possible to re-identify automatically with such models, then using them during sanitization could lead to more robust and future-proof sanitization. One could use these models during sanitization to verify whether certain documents being leaked/released could lead to a higher risk of private data being re-identified.

 The three datasets employed in the experiments consist of either publicly available data (Wikipedia, court rulings from the ECHR) or synthetic documents (clinical notes).

\bibliography{anthology,custom}

\appendix

\section{Description of NER Categories}

\subsection{Wikipeadia Biographies} \label{sec:wiki_ner}

    These descriptions come directly from Spacy.\footnote{https://spacy.io/}

    \begin{description}
        \item \textbf{CARDINAL} Numerals that do not fall under another type
        \item \textbf{DATE} Absolute or relative dates or periods
        \item \textbf{EVENT} Named hurricanes, battles, wars, sports events, etc.
        \item \textbf{FAC} Buildings, airports, highways, bridges, etc.
        \item \textbf{GPE} Countries, cities, states
        \item \textbf{LANGUAGE} Any named language
        \item \textbf{LAW} Named documents made into laws.
        \item \textbf{LOC} Non-GPE locations, mountain ranges, bodies of water
        \item \textbf{MONEY} Monetary values, including unit
        \item \textbf{NORP} Nationalities or religious or political groups
        \item \textbf{ORDINAL} ``first'', ``second'', etc.
        \item \textbf{ORG} Companies, agencies, institutions, etc.
        \item \textbf{PERCENT} Percentage, including ``\%''
        \item \textbf{PERSON} People, including fictional
        \item \textbf{PRODUCT} Objects, vehicles, foods, etc. (not services)
        \item \textbf{QUANTITY} Measurements, as of weight or distance
        \item \textbf{TIME} Times smaller than a day
        \item \textbf{WORK\_OF\_ART} Titles of books, songs, etc.
    \end{description}

\subsection{TAB} \label{sec:tab_ner}

    These descriptions come from \citep{benchmark}.

    \begin{description}
        \item \textbf{CODE} Numbers and identification codes, such as social security numbers, phone numbers, passport numbers, or license plates.
        \item \textbf{DATETIME} Description of a specific date, time, or duration.
        \item \textbf{DEM} Demographic attributes of a person, such as native language, descent, heritage, ethnicity, job titles, ranks, education, physical descriptions, diagnosis, birthmarks, and ages.
        \item \textbf{LOC} Places and locations, such as cities, areas, countries, addresses, named infrastructures, etc.
        \item \textbf{MISC} All other types of personal information associated (directly or indirectly) with an individual and that does not belong to the categories above.
        \item \textbf{ORG} Names of organizations, such as public and private companies, schools, universities, public institutions, prisons, healthcare institutions, non-governmental organizations, churches, etc.
        \item \textbf{PERSON} Names of people, including nicknames/aliases, usernames, and initials.
        \item \textbf{QUANTITY} Description of a meaningful quantity, e.g., percentages or monetary values.
    \end{description}

\onecolumn
\section{GLM Results} \label{app:glm_recall}

\subsection{Wikipedia}

    \cref{tab:glm_wiki_re_id_em} contains the detailed exact match results of the GLM infilling model on Wikipedia Biographies.\cref{tab:app_glm_wiki_re_id_recall} contains the detailed token recall results of the GLM infilling model on Wikipedia Biographies.

\begin{table*}[ht]
    \small
     \centering
     \begin{tabular}{lccccccc@{}} 
          \toprule  
          \multicolumn{1}{@{}c}{\textbf{NER Category}}&  \textbf{No retrieval}&  \multicolumn{2}{c}{\textbf{Not Biographies}}&  \multicolumn{2}{c}{\textbf{All but not original}}& \multicolumn{2}{c}{\textbf{All}}\\
          \cmidrule(lr){3-4}\cmidrule(lr){5-6}\cmidrule(l){7-8}
          & & k=1 & k=2 & k=1 & k=2 & k=1 & k=2 \\
          \midrule
        \multicolumn{1}{@{}l}{\textsc{GLM}} & \hphantom{0}\textbf{6.26} & \hphantom{0}\textbf{7.63} & \hphantom{0}\textbf{7.71} & \hphantom{0}\textbf{9.56} & \hphantom{0}\textbf{9.77} & \textbf{80.08} & \textbf{78.99} \\
        CARDINAL & 28.22 & 28.91 & 28.96 & 30.98 & 29.57 & 83.70 & 81.35 \\
        DATE & \hphantom{0}3.87 & \hphantom{0}4.88 & \hphantom{0}5.10 & \hphantom{0}6.38 & \hphantom{0}6.99 & 78.94 & 79.10 \\
        EVENT & \hphantom{0}6.31 & \hphantom{0}7.71 & \hphantom{0}7.63 & 16.95 & 18.05 & 81.29 & 79.84 \\
        FAC & \hphantom{0}1.71 & \hphantom{0}2.94 & \hphantom{0}2.13 & \hphantom{0}4.41 & \hphantom{0}3.62 & 84.45 & 83.83 \\
        GPE & \hphantom{0}5.24 & \hphantom{0}6.50 & \hphantom{0}6.72 & \hphantom{0}9.49 & \hphantom{0}9.93 & 85.51 & 83.14 \\
        LANGUAGE & 14.84 & 14.65 & 15.75 & 20.15 & 18.68 & 92.31 & 89.35 \\
        LAW & \hphantom{0}0.00 & \hphantom{0}0.00 & \hphantom{0}0.00 & 14.29 & 14.29 & 90.47 & 90.47 \\
        LOC & 10.27 & 16.44 & 16.21 & 15.52 & 15.43 & 90.64 & 90.85 \\
        MONEY & \hphantom{0}4.17 & \hphantom{0}5.56 & \hphantom{0}4.86 & \hphantom{0}4.86 & \hphantom{0}6.25 & 83.33 & 84.72 \\
        NORP & 18.13 & 22.08 & 22.14 & 26.86 & 26.63 & 90.90 & 88.54 \\
        ORDINAL & 50.98 & 50.33 & 49.51 & 54.01 & 54.97 & 87.90 & 87.18 \\
        ORG & \hphantom{0}3.35 & \hphantom{0}5.56 & \hphantom{0}5.68 & \hphantom{0}7.51 & \hphantom{0}7.64 & 80.39 & 79.23 \\
        PERCENT & \hphantom{0}0.00 & \hphantom{0}1.96 & \hphantom{0}3.92 & \hphantom{0}9.80 & \hphantom{0}9.80 & 82.35 & 80.39 \\
        PERSON & \hphantom{0}0.82 & \hphantom{0}1.75 & \hphantom{0}1.94 & \hphantom{0}2.54 & \hphantom{0}2.78 & 74.49 & 74.30 \\
        PRODUCT & \hphantom{0}3.60 & \hphantom{0}4.47 & \hphantom{0}4.46 & \hphantom{0}3.78 & \hphantom{0}5.50 & 85.23 & 83.16 \\
        QUANTITY & \hphantom{0}4.24 & \hphantom{0}2.26 & \hphantom{0}3.39 & \hphantom{0}5.08 & \hphantom{0}5.15 & 75.14 & 63.84 \\
        TIME & \hphantom{0}4.17 & \hphantom{0}3.33 & \hphantom{0}1.67 & \hphantom{0}2.78 & \hphantom{0}2.78 & 78.89 & 73.33 \\
        WORK\_OF\_ART & \hphantom{0}1.93 & \hphantom{0}3.07 & \hphantom{0}2.38 & \hphantom{0}3.79 & \hphantom{0}4.26 & 77.54 & 77.06 \\
         \bottomrule
     \end{tabular}
     \caption{Exact Match of the GLM re-identifier at multiple background knowledge levels and numbers of retrieval texts on the Wikipedia biographies. The overall results are on the same lines as the model name and are bolded. All results are the average of 3 runs, the standard deviation is less than 1\%. Description of categories can be found in \cref{sec:wiki_ner}.}
     \label{tab:glm_wiki_re_id_em} \vspace{-2mm}
 \end{table*}

\begin{table*}[h]
    \small
     \centering
     \begin{tabular}{lccccccc@{}} 
          \toprule  
          \multicolumn{1}{@{}c}{\textbf{NER Category}}&  \textbf{No retrieval}&  \multicolumn{2}{c}{\textbf{Not Biographies}}&  \multicolumn{2}{c}{\textbf{All but not original}}& \multicolumn{2}{c}{\textbf{All}}\\
          \cmidrule(lr){3-4}\cmidrule(lr){5-6}\cmidrule(l){7-8}
          & & k=1 & k=2 & k=1 & k=2 & k=1 & k=2 \\
          \midrule
          \multicolumn{1}{@{}l}{\textsc{GLM}} & \textbf{13.45} & \textbf{14.73} & \textbf{14.99} & \textbf{16.17} & \textbf{16.53} & \textbf{79.76} & \textbf{82.94} \\
          CARDINAL & 28.75 & 30.74 & 30.93 & 31.71 & 32.15 & 83.94 & 87.52 \\
          DATE & 19.27 & 20.72 & 21.04 & 22.29 & 22.84 & 77.43 & 80.93 \\
          EVENT & 31.63 & 34.11 & 34.01 & 36.45 & 36.15 & 86.96 & 89.69 \\
          FAC & 11.08 & 11.73 & 11.75 & 13.89 & 13.81 & 80.63 & 86.67 \\
          GPE & 6.41 & 7.30 & 7.56 & 9.79 & 10.04 & 79.21 & 82.74 \\
          LANGUAGE & 27.21 & 23.81 & 25.09 & 33.45 & 30.55 & 83.73 & 89.94 \\
          LAW & 32.82 & 34.10 & 39.47 & 35.19 & 40.05 & 91.34 & 94.55 \\
          LOC & 10.74 & 11.02 & 10.56 & 10.34 & 12.45 & 82.12 & 84.43 \\
          MONEY & 24.94 & 26.07 & 27.97 & 29.10 & 29.68 & 87.24 & 91.26 \\
          NORP & 17.16 & 19.62 & 20.12 & 23.08 & 24.67 & 84.69 & 87.45 \\
          ORDINAL & 43.99 & 44.54 & 46.27 & 46.97 & 45.94 & 84.81 & 86.17 \\
          ORG & 19.08 & 20.63 & 21.02 & 21.73 & 22.29 & 80.37 & 83.71 \\
          PERCENT & 23.86 & 29.52 & 28.14 & 31.34 & 32.27 & 92.21 & 90.99 \\
          PERSON & 3.16 & 4.20 & 4.36 & 5.29 & 5.35 & 79.07 & 81.76 \\
          PRODUCT & 8.04 & 10.42 & 9.12 & 10.72 & 12.27 & 79.94 & 86.20 \\
          QUANTITY & 27.00 & 29.67 & 28.10 & 34.67 & 37.96 & 90.16 & 88.31 \\
          TIME & 21.28 & 23.84 & 22.02 & 22.75 & 19.92 & 82.17 & 85.25 \\
          WORK\_OF\_ART & 12.50 & 13.35 & 13.81 & 14.81 & 14.99 & 77.89 & 81.34 \\
          \bottomrule
     \end{tabular}
     \caption{Token recall of the GLM re-identifier at various levels of background knowledge and numbers of retrieval texts on the Wikipedia biographies. The overall results are on the same lines as the model name and are bolded. All results are the average of 3 runs, the standard deviation is less than 1\%. Description of categories can be found in \cref{sec:wiki_ner}.}
     \label{tab:app_glm_wiki_re_id_recall} \vspace{-2mm}
 \end{table*}

 \subsection{TAB}

 \cref{tab:glm_tab_re_id} contains the detailed results of the GLM infilling model on TAB.

 \begin{table*}[ht]
     \centering
     \small
     \begin{tabular}{@{}lr@{ / }rr@{ / }lr@{ / }lr@{ / }r@{}} 
          \toprule  
          \textbf{Entity Category} &  \multicolumn{2}{c}{\clap{\textbf{No retrieval}}} &  \multicolumn{2}{c}{\clap{\textbf{General Knowledge}}} & \multicolumn{2}{c}{\clap{\textbf{All but not original}}} & \multicolumn{2}{c}{\textbf{All}} \\
          \midrule
            CODE & 0.10 & 14.25 & \hskip15pt\relax 1.70 & 14.47\hskip15pt\null & \hskip15pt\relax 1.31 & 32.21\hskip15pt\null & 18.14 & 46.70 \\
            DATETIME & 0.25 & 5.31 & 13.42 & 22.68 & 14.53 & 24.89 & 79.32 & 83.27 \\
            DEM & 9.04 & 13.98 & 20.18 & 31.56 & 24.64 & 37.21 & 64.56 & 73.58 \\
            LOC & 5.13 & 5.47 & 19.82 & 21.69 & 21.31 & 24.01 & 75.44 & 81.29 \\
            MISC & 0.00 & 5.52 & 5.37 & 22.49 & 14.84 & 37.40 & 56.14 & 76.87 \\
            ORG & 0.00 & 8.55 & 16.03 & 34.46 & 20.82 & 39.61 & 76.65 & 88.80 \\
            PERSON & 0.04 & 4.01 & 1.31 & 10.20 & 10.14 & 27.63 & 40.28 & 57.96 \\
            QUANTITY & 0.00 & 13.33 & 13.42 & 34.85 & 11.74 & 38.77 & 63.01 & 86.20 \\
            \midrule
            DIRECT & 0.00 & 6.61 & 0.82 & 7.12 & 0.81 & 15.23 & 28.60 & 47.63 \\
            QUASI & 0.90 & 6.23 & 12.08 & 22.63 & 15.40 & 30.54 & 68.93 & 77.82 \\
            \midrule
            AVERAGE & 0.84 & 6.26 & 11.27 & 21.35 & 14.32 & 29.08 & 66.04 & 75.13 \\
          \bottomrule
     \end{tabular}
     \caption{Results of the GLM infiller at multiple background knowledge levels on TAB. The first result represents exact match performance and the second is token recall. Description of categories can be found in \cref{sec:tab_ner}. All results are the average of 3 runs, the standard deviation is less than 1\%.}
     \label{tab:glm_tab_re_id}
\end{table*}

 \subsection{Clinical Cases}

 \cref{tab:glm_clinical_re_id} contains the detailed results of the GLM infilling model on the Clinical Notes dataset.

 \begin{table*}[h]
     \centering
     \small
     \begin{tabular}{@{}lr@{ / }rr@{ / }lr@{ / }lr@{ / }r@{}} 
          \toprule  
          \textbf{Entity Category} &  \multicolumn{2}{c}{\clap{\textbf{No retrieval}}} &  \multicolumn{2}{c}{\clap{\textbf{General Knowledge}}} & \multicolumn{2}{c}{\clap{\textbf{All but not original}}} & \multicolumn{2}{c}{\textbf{All}} \\
          \midrule
            CARDINAL & 48.16 & 44.86 & \hskip15pt\relax 47.73 & 44.34\hskip15pt\null & \hskip15pt\relax 58.39 & 56.46\hskip15pt\null & 91.64 & 90.04 \\
            DATE & 9.80 & 37.14 & 9.66 & 33.71 & 29.81 & 61.98 & 91.40 & 95.24 \\
            GPE & 0.00 & 0.98 & 0.64 & 0.87 & 62.82 & 60.49 & 98.57 & 97.82 \\
            LAW & 0.00 & 0.00 & 0.00 & 0.00 & 100.00 & 100.00 & 100.00 & 100.00 \\
            NORP & 4.97 & 6.35 & 8.51 & 9.42 & 65.96 & 65.37 & 98.58 & 98.91 \\
            ORDINAL & 30.16 & 26.23 & 30.16 & 27.91 & 33.33 & 27.22 & 92.07 & 92.07 \\
            ORG & 0.00 & 3.68 & 0.00 & 3.04 & 62.59 & 62.25 & 51.02 & 73.88 \\
            PERCENT & 20.00 & 55.15 & 40.00 & 56.43 & 53.33 & 83.33 & 100.00 & 100.00 \\
            PERSON & 0.20 & 3.27 & 0.47 & 7.12 & 14.84 & 31.88 & 83.54 & 87.47 \\
            PRODUCT & 2.08 & 26.78 & 6.25 & 36.01 & 23.75 & 58.18 & 79.58 & 92.16 \\
            QUANTITY & 0.45 & 11.72 & 9.31 & 20.04 & 34.53 & 52.71 & 91.89 & 93.87 \\
            TIME & 23.14 & 26.82 & 23.78 & 26.35 & 59.66 & 69.73 & 94.90 & 96.06 \\
            \midrule
            AVERAGE & 18.31 & 26.71 & 18.92 & 26.36 & 42.31 & 55.40 & 90.87 & 92.68 \\
          \bottomrule
     \end{tabular}
     \caption{Results of the GLM infiller at multiple background knowledge levels on the Clinical notes dataset. The first result represents exact match performance and the second is token recall. All results are the average of 3 runs, the standard deviation is less than 1\%. Description of categories can be found in \cref{sec:wiki_ner}.}
     \label{tab:glm_clinical_re_id} \vspace{-2mm}
 \end{table*}

\section{Mistral-Nemo-Instruct Results} \label{app:mistral}

\subsection{TAB}

 \cref{tab:mistral_tab_re_id} contains the detailed results of the Mistral infilling model on TAB.

    \begin{table*}[h]
     \centering
     \small
     \begin{tabular}{@{}lr@{ / }rr@{ / }lr@{ / }lr@{ / }r@{}} 
          \toprule  
          \textbf{Entity Category} &  \multicolumn{2}{c}{\clap{\textbf{No retrieval}}} &  \multicolumn{2}{c}{\clap{\textbf{General Knowledge}}} & \multicolumn{2}{c}{\clap{\textbf{All but not original}}} & \multicolumn{2}{c}{\textbf{All}} \\
          \midrule
            CODE & 0.00 & 43.32 & \hskip15pt\relax 6.19 & 56.22\hskip15pt\null & \hskip15pt\relax 2.22 & 64.21\hskip15pt\null & 13.98 & 71.34 \\
            DATETIME & 0.01 & 36.82 & 10.44 & 63.48 & 10.75 & 65.55 & 45.52 & 83.72 \\
            DEM & 11.92 & 16.36 & 21.93 & 27.33 & 24.27 & 30.29 & 47.66 & 52.21 \\
            LOC & 3.55 & 5.81 & 25.63 & 25.46 & 26.00 & 24.68 & 54.24 & 53.56 \\
            MISC & 0.00 & 8.54 & 7.60 & 21.49 & 12.10 & 24.50 & 30.92 & 57.52 \\
            ORG & 1.38 & 4.21 & 17.75 & 30.16 & 20.07 & 21.75 & 45.96 & 63.26 \\
            PERSON & 0.00 & 0.68 & 1.05 & 9.91 & 3.72 & 16.77 & 14.30 & 33.57 \\
            QUANTITY & 0.00 & 46.48 & 3.58 & 53.01 & 2.82 & 55.60 & 15.98 & 76.38 \\
            \midrule
            DIRECT & 0.00 & 15.84 & 5.71 & 36.36 & 0.90 & 33.96 & 15.39 & 48.86 \\
            QUASI & 0.98 & 25.90 & 10.97 & 48.49 & 12.12 & 49.25 & 39.03 & 72.21 \\
            \midrule
            AVERAGE & 0.91 & 25.36 & 10.59 & 47.43 & 11.00 & 47.98 & 37.34 & 70.29 \\
          \bottomrule
     \end{tabular}
     \caption{Results of the Mistral infiller at multiple background knowledge levels on TAB. The first result represents exact match performance and the second is token recall. All results are the average of 3 runs, the standard deviation is less than 1\%. Description of categories can be found in \cref{sec:tab_ner}.}
     \label{tab:mistral_tab_re_id} \vspace{-2mm}
 \end{table*}

\subsection{Clinical Notes}

 \cref{tab:mistral_clinical_re_id} contains the detailed results of the Mistral infilling model on the Clinical Notes dataset.

 \begin{table*}[h]
     \centering
     \small
     \begin{tabular}{@{}lr@{ / }rr@{ / }lr@{ / }lr@{ / }r@{}} 
          \toprule  
          \textbf{Entity Category \ } &  \multicolumn{2}{c}{\textbf{No retrieval \ }} &  \multicolumn{2}{c}{\textbf{General Knowledge \ }} & \multicolumn{2}{c}{\textbf{All but not original \ }}& \multicolumn{2}{c}{\textbf{All}} \\
          \midrule
            CARDINAL & 4.54 & 4.46 & 6.03 & 5.98 & 10.72 & 9.37 & 15.00 & 11.65 \\
            DATE & 1.96 & 42.27 & 2.46 & 39.64 & 17.03 & 62.43 & 30.20 & 80.17 \\
            GPE & 13.25 & 7.84 & 0.93 & 0.17 & 55.41 & 44.77 & 59.26 & 56.59 \\
            LAW & 100.00 & 100.00 & 100.00 & 100.00 & 100.00 & 100.00 & 100.00 & 100.00 \\
            NORP & 2.13 & 2.30 & 11.35 & 14.81 & 16.31 & 13.54 & 21.28 & 20.10 \\
            ORDINAL & 4.76 & 2.98 & 3.17 & 1.17 & 22.22 & 12.65 & 57.14 & 43.51 \\
            ORG & 71.43 & 79.75 & 72.79 & 54.68 & 73.47 & 72.86 & 91.16 & 93.83 \\
            PERCENT & 0.00 & 48.08 & 0.00 & 57.14 & 0.00 & 61.90 & 0.00 & 100.00 \\
            PERSON & 0.40 & 10.51 & 0.88 & 9.20 & 19.50 & 25.80 & 44.53 & 50.91 \\
            PRODUCT & 5.00 & 32.66 & 17.08 & 57.08 & 10.00 & 49.93 & 27.92 & 81.55 \\
            QUANTITY & 0.00 & 31.58 & 0.75 & 45.40 & 8.41 & 68.20 & 13.21 & 88.01 \\
            TIME & 3.50 & 43.34 & 7.64 & 41.29 & 10.40 & 61.27 & 14.86 & 76.22 \\
            \midrule
            AVERAGE & 4.76 & 23.46 & 4.52 & 17.08 & 19.88 & 40.80 & 30.19 & 57.85 \\
          \bottomrule
     \end{tabular}
     \caption{Results of the Mistral infiller at multiple background knowledge levels on the Clinical notes dataset. The first result represents exact match performance and the second is token recall. All results are the average of 3 runs, the standard deviation is less than 1\%. Description of categories can be found in \cref{sec:wiki_ner}.}
     \label{tab:mistral_clinical_re_id} \vspace{-2mm}
 \end{table*}

\section{Prompts}\label{app:prompts}

\subsection{Prompts to generate articles}

\noindent

\subsubsection{The Guardian article}

Assume you are an investigative journalist for The Guardian, and in charge of covering human rights abuses. You have just been presented the following court ruling from the European Court of Human Rights: 

\{CASE\}  

Now write a news article that covers the key facts of the case, the outcome of the ruling, and what it may mean for the protection of human rights in Europe.

\subsubsection{Blog Post}

Assume you are an avid blogger who reports on human rights abuses. You have just been presented the following court ruling from the European Court of Human Rights:

\{CASE\}

Now write a blog post that covers the key facts of the case, the outcome of the ruling, and what it may mean for the protection of human rights in Europe.

\subsubsection{Court Report}

Assume you are a court reporter which must give a detailed account of human rights abuse cases. You have just been presented the following court ruling from the European Court of Human Rights:

\{CASE\}

Now write a detailed court report that covers the key facts of the case and the outcome of the ruling.

\subsection{Prompts to infill}

\subsubsection{No retrieval}

Re-identify the fill in the blank (marked with [MASK]) in the text below, only give the value of the [MASK], do not add extra text, give explanations, or output the blank token [MASK]:

\{text\}

Answer:

\subsubsection{Retrieval}

Given the following passages:

\{retrieved\}

Re-identify the fill in the blank (marked with [MASK]) in the text below, only give the value of the [MASK], do not add extra text, give explanations, or output the blank token [MASK]:

\{text\}

Answer:

\section{Re-identifications}\label{app:re-identification}

All the following will be on TAB at various levels of background knowledge

\subsection{Original}

The case originated in an application (no. \textbf{39958/02}) against the Republic of Poland lodged with the Court under Article 34 of the Convention for the Protection of Human Rights and Fundamental Freedoms (“the Convention”) by a Polish national, \textbf{Mr Dariusz Piątkiewicz} (“the applicant”), on \textbf{14 June 2000}.

The Polish Government were represented by their Agent, \textbf{Mr J. Wołąsiewicz} of the Ministry of Foreign Affairs.

On \textbf{1 September 2006} the President of the Fourth Section of the Court decided to communicate the application to the Government. Under the provisions of Article 29 § 3 of the Convention, it was decided to examine the merits of the application at the same time as its admissibility.

\subsection{GLM}

\subsubsection{No Retrieval (Level 1)}

The case originated in an application (no. \textbf{1948}) against the Republic of Poland lodged with the Court under Article 34 of the Convention for the Protection of Human Rights and Fundamental Freedoms (“the Convention”) by a Polish national, \textbf{Jan Klimczak} (“the applicant”), on \textbf{the same day}.

The Polish Government were represented by their Agent, \textbf{Jan Klimczak} of the Ministry of Foreign Affairs.

On \textbf{the same day} the President of the Fourth Section of the Court decided to communicate the application to the Government. Under the provisions of Article 29 § 3 of the Convention, it was decided to examine the merits of the application at the same time as its admissibility.

\subsubsection{General (Level 2)}

The case originated in an application (no. \textbf{39}) against the Republic of Poland lodged with the Court under Article 34 of the Convention for the Protection of Human Rights and Fundamental Freedoms (“the Convention”) by a Polish national, \textbf{Jan Klimczak} (“the applicant”), on \textbf{9 October 2003}.

The Polish Government were represented by their Agent, \textbf{Mr Krzysztof Drzewicki} of the Ministry of Foreign Affairs.

On \textbf{29 October 2003} the President of the Fourth Section of the Court decided to communicate the application to the Government. Under the provisions of Article 29 § 3 of the Convention, it was decided to examine the merits of the application at the same time as its admissibility.

\subsubsection{No Original (Level 3)}

The case originated in an application (no. \textbf{48247/06}) against the Republic of Poland lodged with the Court under Article 34 of the Convention for the Protection of Human Rights and Fundamental Freedoms (“the Convention”) by a Polish national, \textbf{Mr Stanisław Opałko} (“the applicant”), on \textbf{22 January 2003}.

The Polish Government were represented by their Agent, \textbf{Mr J. Wołąsiewicz} of the Ministry of Foreign Affairs.

On \textbf{19 March 2007} the President of the Fourth Section of the Court decided to communicate the application to the Government. Under the provisions of Article 29 § 3 of the Convention, it was decided to examine the merits of the application at the same time as its admissibility.

\subsubsection{All (Level 4)}

The case originated in an application (no. \textbf{48247/06}) against the Republic of Poland lodged with the Court under Article 34 of the Convention for the Protection of Human Rights and Fundamental Freedoms (“the Convention”) by a Polish national, \textbf{Mr Stanisław Opałko} (“the applicant”), on \textbf{22 January 2003}.

The Polish Government were represented by their Agent, \textbf{Mr J. Wołąsiewicz} of the Ministry of Foreign Affairs.

On \textbf{1 September 2006} the President of the Fourth Section of the Court decided to communicate the application to the Government. Under the provisions of Article 29 § 3 of the Convention, it was decided to examine the merits of the application at the same time as its admissibility.

\subsection{Mistral}

\subsubsection{No Retrieval (Level 1)}

The case originated in an application (no. \textbf{1}) against the Republic of Poland lodged with the Court under Article 34 of the Convention for the Protection of Human Rights and Fundamental Freedoms (“the Convention”) by a Polish national, \textbf{Polish} (“the applicant”), on \textbf{Poland}.

The Polish Government were represented by their Agent, \textbf{Polish} of the Ministry of Foreign Affairs.

On \textbf{Polish} the President of the Fourth Section of the Court decided to communicate the application to the Government. Under the provisions of Article 29 § 3 of the Convention, it was decided to examine the merits of the application at the same time as its admissibility.

\subsubsection{General (Level 2)}

The case originated in an application (no. \textbf{38426/03}) against the Republic of Poland lodged with the Court under Article 34 of the Convention for the Protection of Human Rights and Fundamental Freedoms (“the Convention”) by a Polish national, \textbf{Lidia Nowak} (“the applicant”), on \textbf{9 October 2003}.

The Polish Government were represented by their Agent, \textbf{Krzysztof Drzewicki} of the Ministry of Foreign Affairs.

On \textbf{12 October 2022} the President of the Fourth Section of the Court decided to communicate the application to the Government. Under the provisions of Article 29 § 3 of the Convention, it was decided to examine the merits of the application at the same time as its admissibility.

\subsubsection{No Original (Level 3)}

The case originated in an application (no. \textbf{473/07}) against the Republic of Poland lodged with the Court under Article 34 of the Convention for the Protection of Human Rights and Fundamental Freedoms (“the Convention”) by a Polish national, \textbf{Stanisław Opałko} (“the applicant”), on \textbf{22 January 2003}.

The Polish Government were represented by their Agent, \textbf{J. Wołąsiewicz} of the Ministry of Foreign Affairs.

On \textbf{22 January 2003} the President of the Fourth Section of the Court decided to communicate the application to the Government. Under the provisions of Article 29 § 3 of the Convention, it was decided to examine the merits of the application at the same time as its admissibility.

\subsubsection{All (Level 4)}

The case originated in an application (no. \textbf{473/07}) against the Republic of Poland lodged with the Court under Article 34 of the Convention for the Protection of Human Rights and Fundamental Freedoms (“the Convention”) by a Polish national, \textbf{Stanisław Opałko} (“the applicant”), on \textbf{22 January 2003}.

The Polish Government were represented by their Agent, \textbf{J. Wołąsiewicz} of the Ministry of Foreign Affairs.

On \textbf{19 March 2007} the President of the Fourth Section of the Court decided to communicate the application to the Government. Under the provisions of Article 29 § 3 of the Convention, it was decided to examine the merits of the application at the same time as its admissibility.

\section{Ranking results}\label{Ranking}

\subsection{Accuracy@1}

\cref{tab:ranking1} shows the accuracy at 1 for the re-identification model.

 \begin{table}[ht]
     \centering
     \small
     \begin{tabular}{@{}lccccc@{}} 
          \toprule  
          \textbf{Dataset} & \textbf{Masked} & \textbf{L1} & \textbf{L2} & \textbf{L3} & \textbf{L4} \\
          \midrule
          \textsc{GLM} & & & & & \\
          \hspace{.3em}TAB & 5.5 & 5.5 & 7.1 & 4.7 & 37.0 \\
          \hspace{.3em}Clinical & 46.3 & 48.7 & 49.3 & 62.8 & 82.9 \\[0.5em]
          \textsc{Mistral} & & & & & \\
          \hspace{.3em}TAB & 5.5 & 3.1 & 8.7 & 7.9 & 30.7 \\
          \hspace{.3em}Clinical & 46.3 & 47.7 & 52.7 & 67.4 & 84.2 \\
          \bottomrule
     \end{tabular}
     \caption{Percentage of documents in which the ranking model has the correct person identity as its top prediction.}
     \label{tab:ranking1} \vspace{-2mm}
 \end{table}

\subsection{Accuray@5}

\cref{tab:ranking5} shows the accuracy at 5 for the re-identification model.

\begin{table}[ht]
     \centering
     \small
     \begin{tabular}{@{}lccccc@{}} 
          \toprule  
          \textbf{Dataset} & \textbf{Masked} & \textbf{L1} & \textbf{L2} & \textbf{L3} & \textbf{L4} \\
          \midrule
          \textsc{GLM} & & & & & \\
          \hspace{.3em}TAB & 16.5 & 20.5 & 18.9 & 16.5 & 54.3 \\
          \hspace{.3em}Clinical & 50.3 & 54.7 & 55.7 & 68.8 & 90.9 \\[0.5em]
          \textsc{Mistral} & & & & & \\
          \hspace{.3em}TAB & 16.5 & 17.3 & 22.8 & 24.4 & 50.4 \\
          \hspace{.3em}Clinical & 50.3 & 56.4 & 59.7 & 74.5 & 92.3 \\
          \bottomrule
     \end{tabular}
     \caption{Percentage of re-identified documents in which the correct identity is found in the top-5 predictions.}
     \label{tab:ranking5} \vspace{-2mm}
 \end{table}




\subsection{MRR}

\cref{tab:rankingmrr} shows the accuracy at MRR for the re-identification model.

\begin{table}[ht]
     \centering
     \small
     \begin{tabular}{@{}lccccc@{}} 
          \toprule  
          \textbf{Dataset} & \textbf{Masked} & \textbf{L1} & \textbf{L2} & \textbf{L3} & \textbf{L4} \\
          \midrule
          \textsc{GLM} & & & & & \\
          \hspace{.3em}TAB & 0.135 & 0.144 & 0.154 & 0.138 & 0.473 \\
          \hspace{.3em}Clinical & 0.496 & 0.525 & 0.530 & 0.662 & 0.862 \\[0.5em]
          \textsc{Mistral} & & & & & \\
          \hspace{.3em}TAB & 0.135 & 0.126 & 0.175 & 0.180 & 0.416 \\
          \hspace{.3em}Clinical & 0.496 & 0.520 & 0.566 & 0.708 & 0.871 \\
          \bottomrule
     \end{tabular}
     \caption{The Mean Reciprocal Rank of the correct identity.}
     \label{tab:rankingmrr} \vspace{-2mm}
 \end{table}

\end{document}